\documentclass[letterpaper, 10 pt, conference]{ieeeconf}  

\IEEEoverridecommandlockouts

\usepackage[inkscapelatex=false]{svg}
\usepackage{stfloats} 
\usepackage{amsmath,amsfonts}
\usepackage{algorithmic}
\usepackage{algorithm}
\usepackage{array}
\usepackage[caption=false,font=footnotesize,labelfont=sf,textfont=sf]{subfig}

\usepackage{textcomp}
\usepackage{stfloats}
\usepackage{url}
\usepackage{verbatim}
\usepackage{graphicx}
\usepackage{cite}
\usepackage{soul}
\usepackage{adjustbox}
\usepackage{booktabs}
\usepackage{capt-of}
\usepackage{todonotes}
\usepackage{soul}
\usepackage{multirow}
\usepackage{balance}
\usepackage[dvipsnames]{xcolor}

\title{\LARGE \bf

Revisiting Direct Control: A Field Evaluation of Sensor Fusion Efficacy in Teleoperating Full-Scale Autonomous Vehicles via Bird's Eye View}

\author{Stanisław Kuczma$^{1}$, Sławomir K. Tadeja$^{2}$, Michał R. Nowicki$^{1}$
\thanks{S. Kuczma and M.R. Nowicki are with the $^{1}$Poznań University of Technology., S.K. Tadeja is with the $^{2}$Massachusetts Institute of Technology. Contact: \texttt{stanislaw.kuczma@put.poznan.pl} This paper is based on the thesis by S. Kuczma.
}
\thanks{Approval for experimental procedures was granted by the Poznań University of Technology Ethics Committee (Approval No. 006/2025 29.05.2025).
}
\thanks{This research was supported by OMAG Sp.~z~o.o.\ under the project ``Wielozadaniowa, bezza\l{}ogowa platforma terenowa'', co-financed by the European Funds for Ma\l{}opolska 2021--2027 programme (Agreement No.~FEMP.01.01-IP.01-0270/23-00).
}
\thanks{Work by M. R. Nowicki was financed by Poznan University of Technology internal grant 0214/SBAD/0256.}
}

\begin{document}

\maketitle
\thispagestyle{empty}
\pagestyle{empty}


\begin{abstract}

In autonomous driving, robot teleoperation serves as a critical safety fallback mechanism, requiring operators to assume direct control in the event of system failure. While recent research has pivoted toward high-level supervisory control or simulation-based studies, these approaches often fail to capture the physical dynamics and sensory demands of driving a full-scale vehicle. To address this gap, we evaluated the impact of spatial data representation on operator performance using a real car-sized vehicle in a field experiment with $N=18$ participants. We compared three visualization interfaces: (i) a standard camera feed baseline, with two bird’s-eye views (BEV) from (ii) a raw LiDAR point cloud, and (iii) a sensor fusion of LiDAR and camera data streams. Conversely to prior passive-observation studies suggesting that sensor-fusion BEV may hinder awareness, our active-control results demonstrate that such representation significantly outperforms the standard LiDAR BEV in perceived usability ($p\leq0.0001$) and cognitive task load ($p=0.01$, NASA-TLX score: $34.58/100$) while also being significantly better than the Camera-only view ($p=0.043$). The fused interface also led to statistically significant reductions in task completion time ($p=0.0461$) and superior usability (SUS: $77.9/100$) compared to those of the Camera-only and LiDAR-only views. By synthesizing objective performance data with comprehensive subjective assessments, we demonstrate that BEV fusion maximizes safety margins while minimizing the operator’s mental demand. These findings challenge existing teleoperation conventions and provide an empirically validated approach for designing robust remote piloting systems.

\end{abstract}


%

\section{Introduction}
Self-driving vehicles are rapidly redefining mobility, with industry leaders such as Waymo and Tesla deploying autonomous technologies in public spaces. While such automated driving systems have reached series-level implementation, their operational design domains~\cite{odd} remain restricted to specific environments and low speeds. Consequently, teleoperation has emerged as a critical bridge technology and a vital safety fallback, allowing remote human operators to assume control when the autonomous system encounters a scenario beyond its existing capabilities~\cite{failure}.

\begin{figure}[!th]
\centering
\includegraphics[width=1.0\linewidth]{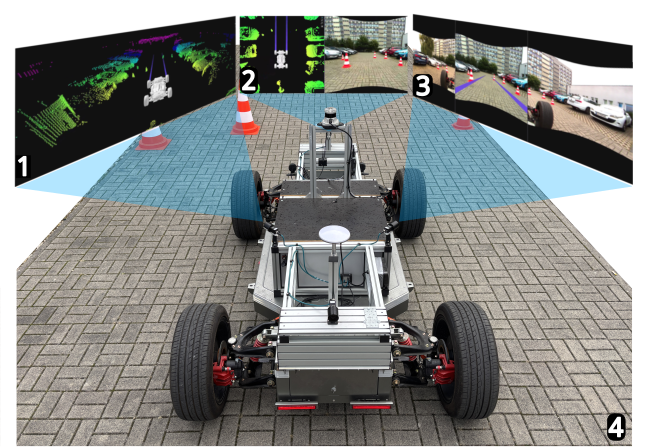}
\caption{Despite the popularity of the bird’s-eye view for machine learning-based perception, robot teleoperation is usually performed with a camera video stream. We asked 18 participants to teleoperate a car-sized vehicle on the (4) obstacle course with (1--3) varying visualization views to quantify their impact on the operator's performance. We collected both subjective usability and workload data, and objective performance metrics, including driving time, number of collisions, mean trajectory deviation, and stopping accuracy, to judge each validated visualization view.}
\label{fig:catchy}
\vspace{-2mm}
\end{figure}

At the same time, consumer preferences underscore the necessity of human-in-the-loop systems. A recent survey~\cite{Heineke_2024} found that 58\% of respondents prefer remote human control over full autonomy for navigating unexpected or ethically complex situations~\cite{autonomous24}. However, the transition to remote control introduces a significant \textit{physical disconnection} between the operator and the vehicle. Teleoperators must contend with transmission latency and restricted field of view (FoV), lacking the ability to sense vehicle forces~\cite{being} or hear~\cite{hearing} their immediate surroundings. Effective data visualization is therefore paramount for bridging this sensory gap and enhancing operator situational awareness~\cite{autonomousvis}. Conversely, poorly designed interfaces risk overloading operators with irrelevant or unclear information, eroding trust in autonomy, and increasing the likelihood of errors \cite{41}.

A critical disparity exists in current literature, i.e. while user interface designs are becoming more sophisticated, many evaluations rely on static clickable prototypes (mock-ups) ~\cite{62,51}, miniature, non-real-life-size robotic platforms~\cite{50,66,53}, and simulations~\cite{65, 68}. These low-fidelity environments often fail to replicate the complex motion inertia and sensory dynamics inherent in full-scale vehicle control~\cite{67}. Furthermore, while the bird’s-eye view (BEV) has become the state-of-the-art representation for machine learning-based perception tasks~\cite{bevreview}, it can be characterized as an inferior or overwhelming visualization for human operators compared to standard camera feeds~\cite{46}.

This paper addresses these gaps by presenting a comprehensive comparative analysis of sensor data representations—ranging from standard perspective RGB video streams to Camera-LiDAR BEV fusion (see Fig.~\ref{fig:catchy}.1-3)—conducted on a full-scale Pixloop vehicle platform (see Fig.~\ref{fig:catchy}.4). Through a field experiment with human operators ($N=18$) driving a real vehicle, we determined which modality best optimized operator performance under real vehicle dynamics.

In summary, we contribute the following:
\begin{enumerate}
    \item The design of a modular teleoperation interface that leverages unified Camera-LiDAR BEV visualizations to mitigate the sensory limitations of remote control and enhance operator performance in the context of full-scale vehicle dynamics.
    \item An empirical, field-based evaluation showing that findings from low-fidelity setups do not fully scale to real-world dynamics, specifically proving that BEV fusion significantly reduces workload and decreases risk in full-scale operations.
\end{enumerate}

\section{Related Work and Motivation}

The challenge of delivering situational awareness for remote operators remains a critical bottleneck in the deployment of teleoperation systems~\cite{47}. We categorize current research efforts based on their sensory data representations and, crucially, the fidelity of their evaluation environments as summarized in Table~\ref{tab:comparison}.
\color{Black}

\begin{table*}[!ht]
\centering
\caption{Comparison and Review of the existing studies coupling remotely operating vehicles with an experiment in terms of platform type, used display setting, data source, transmission, and its representation, as well as experimental sample size (N) or lack thereof, and use of human control showing limited to now user studies utilizing real vehicles.}
\begin{tabular}{lllllll}
\toprule
\textbf{Author} & \textbf{Platform} & \textbf{Display Interface} & \textbf{Data Trans.} & \textbf{Representation} & \textbf{N} & \textbf{Control} \\
\midrule
\multicolumn{7}{c}{\textit{Simulation, Prototype \& Crowdsourcing}} \\
\midrule
Boker \& Lanir~\cite{46} & Crowdsourcing & Monitor & N/A & BEV & 150 & No \\
Kremer et al.~\cite{65} & Simulation (CARLA) & Monitor & Reduced & HD Map + Object List & - & Yes \\
Wolf et al.~\cite{66} & Prototype & Monitor & N/A & 2D Video + Icons & 36 & No \\
Hofbauer et al.~\cite{68} & Simulation (CARLA) & Triple Monitor & Adaptive Bitrate & 2D Video & - & Yes \\
Tener \& Lanir~\cite{62} & Simulation & Monitor & N/A & 3rd Person View + 2D Map & 14 & No \\
\midrule
\multicolumn{7}{c}{\textit{Scale Vehicles \& Robots}} \\
\midrule
Hing et al.~\cite{38} & Gantry System & Monitor & Real-time Video & 3rd Person Video & 7 & Yes \\
Keskinpala~\cite{41} & Mobile Robot & Tablet & Wireless & 2D Video & 30 & Yes \\
Livatino et al.~\cite{23} & Mobile Robot & HMD / Monitor & Wireless & Video + 3D Augment. & 12 & Yes \\
Gafert et al.~\cite{53} & Scale Vehicle (1:13) & XR Headset & Wireless & Video + AR Overlays & 16 & Yes \\
Li et al.~\cite{50} & Mobile Robot & HMD & 5G & 3DGS & 24 & Yes \\
Wilder-Smith~\cite{1} & Robot Arm / Quad. & HMD / Monitor & Wireless & NeRF / 3DGS & 20 & No \\
Wolf et al.~\cite{66} & Scale Vehicle (1:10) & Monitor & WiFi & 2D Video + 3D Map & 45 & Yes \\
\midrule
\multicolumn{7}{c}{\textit{Real Vehicles}} \\
\midrule
Gnatzig et al.~\cite{58} & Real Vehicle & Triple Monitor & Cellular & Stitched video stream & - & Yes \\
Sasai et al.~\cite{61} & Real Vehicle & Projector & N/A & Projected Light (AR) & 8 & No \\
Georg \& Diermeyer~\cite{42} & Real Vehicle & HMD / Monitor & Adaptive & 3D Projection & - & Yes \\
Georg et al.~\cite{48} & Real Vehicle & HMD / Monitor & 4G/LTE & 2D Video + 2D Map & 8 & Yes \\
Schimpe et al.~\cite{33} & Real Vehicle & Monitor & Cellular / WiFi & Video + LiDAR Cloud & - & Yes \\
Kerbl et al.~\cite{59} & Real Vehicle & Triple Monitor & Cellular / WiFi & Multi-Cam Video & - & Yes \\
Hark et al.~\cite{64} & Real Vehicle & Monitor & V2X / 5G & Video + Sensor Fusion & - & Yes \\
Tener \& Lanir~\cite{52} & Real Vehicle & Multi-Monitor & Cellular & Multi-Camera Video & 8 & Yes \\
\midrule
\textbf{Ours} & \textbf{Real Vehicle} & \textbf{Monitor} & \textbf{Wireless (Low-Lat)} & \textbf{Camera + LiDAR BEV} & \textbf{18} & \textbf{Yes} \\
\bottomrule
\end{tabular}

\label{tab:comparison}
\end{table*}


\subsection{Camera-Based Feed}

Camera-based video feeds are fundamental for teleoperation and robotics due to the equipment's low cost, wide availability, and versatility~\cite{autonomous24}. In teleoperation, their role is especially central because they capture the environment in a way that closely resembles human vision, making the data intuitive for operators to interpret~\cite{camerateleop}. 

Most teleoperated vehicles employ a forward-facing camera, often complemented by side-mounted cameras to widen the field of view and avoid blind spots~\cite{58, 59}. However, camera performance strongly depends on lighting conditions, degrading in darkness or under sun glare~\cite{59}. Secondly, direct video feeds lack depth, making it difficult for remote operators to correctly estimate the distance to obstacles~\cite{autonomousvis}. Additional challenges include image distortion at the edges and inconsistencies when stitching images from multiple cameras~\cite{52}.

To mitigate these drawbacks, camera feeds can be projected onto a spatial canvas, an approach widespread in virtual reality (VR)~\cite{48}. This enables third-person perspectives such as ``chase view''~\cite{33, 38}, where the operator observes the vehicle from behind or above. Such perspectives improve trajectory stability and spatial awareness, though the camera’s narrow field of view still limits situational awareness and motivates the integration of additional sensors~\cite{38}.

\subsection{LiDAR-Based Feed}


Research has consistently shown that third-person views outperform first-person views in teleoperation \cite{23, 38}, particularly when enriched with 3D LiDAR data. Central to these views is the inclusion of a vehicle 3D model, which clarifies its spatial position relative to obstacles. This principle holds across platforms -- from quadrupeds to robotic arms~\cite{1}, TurtleBots \cite{50}, autonomous vehicles \cite{59}, and tracked robots \cite{23}. 
However, evaluation of this type of platform is mostly done in miniature scale with unlimited data bandwidth (See Table~\ref{tab:comparison}).

An additional common element is the visualization of the vehicle’s planned trajectory \cite{61}. Already established in automotive research \cite{61}, trajectory overlays have been shown to significantly improve operator performance and confidence~\cite{52,33}. Together, these elements form the foundation of most modern teleoperation interfaces, regardless of whether data originates from cameras, LiDAR, or augmented reconstructions such as radiance fields~\cite{50,1}.

\subsection{Multi-Sensory Feeds Fusion}

Another strategy for enhancing situational awareness during teleoperation is to utilize sensor fusion, which combines multiple sensing modalities~\cite{42}. Sensor fusion in teleoperation has been studied for nearly two decades, with early experiments by Keskinpala and Adams~\cite{41} suggesting that combining visual and depth data increased mental workload and reduced performance. However, subsequent research emphasized the value of redundancy and complementary sensing. For instance, a Munich-developed teleoperation system~\cite{58} provided operators with camera, LiDAR, and radar data, paving the way for later open-source frameworks integrated into ROS~\cite{33} and ROS 2~\cite{59}.

BEV is another widely used representation, enabling operators to observe the vehicle within its environment from above~\cite{zhao2024bev}. While this view facilitates sensor fusion in a unified spatial context, it also presents several challenges. Boker and Lanir~\cite{46} found that a bird's-eye view sometimes hindered the detection of essential features.

While the operator may be considered more informed with the increase of provided data about the vehicle's environment, this also increases the risk of cognitive overload~\cite{cognitive_load}. In practice, when an operator becomes mentally overloaded, performance may drop significantly~\cite{50}. Therefore, as suggested by Gibson’s \textit{Ecological Approach to Visual Perception}~\cite{ecological}, the interface should emphasize the direct perception of meaningful information or affordances rather than raw sensor outputs. By presenting information that aligns with how humans naturally perceive their environment, ecological interfaces aim to improve situational awareness while minimizing overload~\cite{39}. This paper aims to verify if the theoretical advantages of BEV vision have real translation into real-world driving, where direct control loop imposes different demands than passive observations.


\subsection{Practical Evaluation and Fidelity Challenges}


While the aforementioned ecological and multi-sensory principles provide a theoretical roadmap, translating them into reliable teleoperation systems requires addressing critical real-world constraints often overlooked in the existing literature. A significant disparity persists between theoretical user interface sophistication and empirical validation (see Tab.~\ref{tab:comparison}). Most current evaluation studies rely on low-fidelity environments, such as static prototypes or miniature robotic platforms. These environments often fail to replicate the complex motion inertia~\cite{66} and sensory ``physical disconnection'' inherent in the control of full-scale vehicles. Furthermore, even when full-size platforms are utilized, formal human-subject evaluations are often omitted in favor of system architecture demonstrations \cite{64}. 

Modern designs like the EDGAR platform~\cite{59} use modular architectures but suggest keeping interfaces fixed to avoid cognitive overload. Our work fills this gap by providing a rigorous evaluation utilizing a full-scale Pixloop vehicle with a robust $N=18$ participant pool. This allows us to empirically verify if Camera-LiDAR BEV fusion improves performance and reduces workload under authentic dynamics—directly challenging conclusions drawn from prior work

\color{Black}

\section{Materials and Methods}

\subsection{System Architecture and Apparatus}
The test vehicle was based on the full-electric Pixloop~\cite{pixloop} chassis equipped with 3 Lucid TRIO23S-CC cameras, each providing an approximate horizontal field of view of 84\textdegree, yielding images of 960x600 pixels at 15 frames per second placed at the height of 0.6m above the ground, aligned with the horizon with minimal downward tilt, and one 360-degree Ouster OS1-128 3D LiDAR sensor, mounted at the height of 1.4m, providing distance to obstacles in 128 vertical layers with reported resolution of 512 beams at 10~Hz and the maximum distance of 200 meters. 
Ground truth trajectory reference was determined with U-blox ZED-F9P Multi-band GNSS working in RTK mode and mounted on the vehicle. 

\begin{figure}[!t]
\centering
\includegraphics[width=\linewidth]{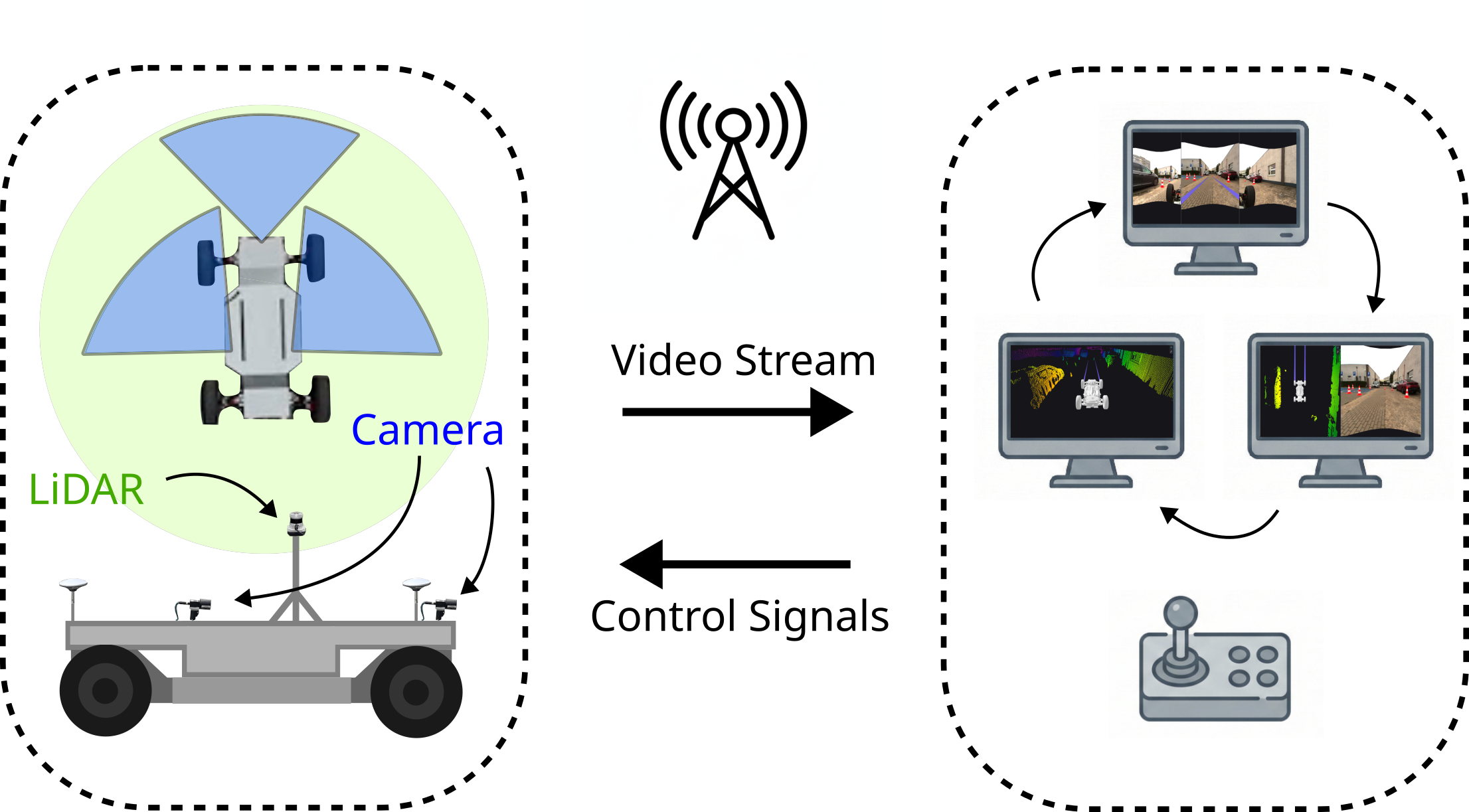}
\caption{The remote operator was positioned at the control station, with operator-vehicle communication split into two independent data streams. The operator's view was obtained with a high-resolution video transmitter, while control signals were used by the official radio controller.}
\label{view_fusion}
\end{figure}

The communication between operator and vehicle was split into two independent streams as presented in Fig.~\ref{view_fusion}.
The video signal to the operator was transmitted using Accsoon HDMI with a latency not exceeding 25 ms \cite{accsoon_data}. However, total end-to-end system latency (including software and display delays) was not formally quantified. The selected setup ensured reliable low-latency data transmission and realistic experimental conditions. 
The entire visualization setup was designed as a complementary feature to the platform’s autonomous capabilities, which remain the primary operational objective. Consequently, the placement of sensors was optimized primarily for autonomous navigation rather than for teleoperation, and the visualization interface was adapted to work with this existing configuration.
The control commands to the vehicle were transmitted using the official, unmodified Pixloop's remote radio controller.

The operators were placed in the control center that was located outside the circuit but on the same parking lot to ensure uninterrupted data transmission and minimal signal latency. It consisted of a Full-HD 24-inch display, Pixloop's official radio controller, and an Accsoon video receiver.

\subsection{Data Representation Layouts}

We developed three different visualization layouts described next. All of them included a projected future trajectory estimation based on the steering angle, providing the operator with predictive cues for vehicle motion.

\begin{figure}[!t]
\centering
\includegraphics[width=\linewidth]{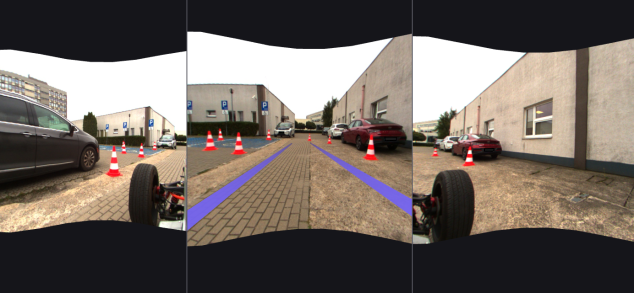}
(A) Camera-based view\\[0.2cm]
\includegraphics[width=\linewidth]{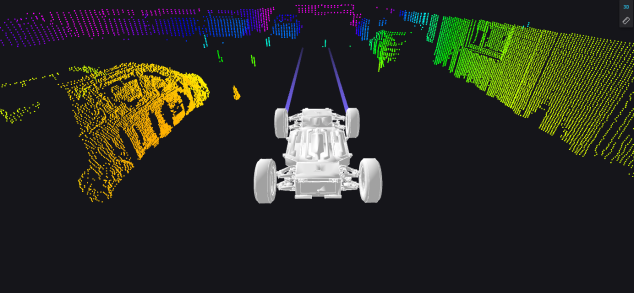}
(B) LiDAR-based BEV\\[0.2cm]
\includegraphics[width=\linewidth]{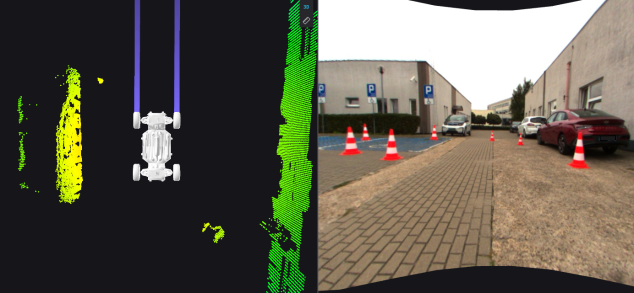}
(C) LiDAR+camera data-fusion BEV\\[0.01cm]
\caption{Operators' views during teleoperation experiment: (A) camera-based, (B) LiDAR-based, and (C) fused camera and LiDAR.}
\label{view_fusion2}
\vspace{-2mm}
\end{figure}

\textbf{Camera-based view} (see Fig.~\ref{view_fusion2}A) was inspired by popular video feed approaches~\cite{58,31,53,52}, featuring a front-facing camera in the center and additional side cameras on supporting panels. 
Multiple cameras capture the vehicle’s surroundings and help the operator estimate relative distances to nearby obstacles~\cite{42}. We also enriched the view with visual aids that show the predicted direction of motion based on the current steering angle~\cite{52,51}. 

\textbf{LiDAR-based view (BEV)} (see Fig.~\ref{view_fusion2}B) presented an isometric perspective designed to enhance depth perception through point cloud visualization. To avoid overloading the operators, the points at ground level were segmented using the Scan Ground Filter module in Autoware Universe\footnote{\url{https://autowarefoundation.github.io/autoware_universe} Accessed 9-2025}. The filter was configured with its default spatial parameters, including a maximum non-ground height threshold of 0.2 meters. The virtual camera was mounted at an elevation of 7 meters above the vehicle, facing forward. Its optical axis was inclined downward by 30 degrees relative to the horizontal plane. Color grading provided by Foxglove was applied to the LiDAR points to indicate distances from 1 to 20 meters, to align with the designated operational focus area.

\textbf{LiDAR+camera data-fusion view (BEV)} (see Fig.~\ref{view_fusion2}C) featured two independent parts: front-facing video stream and a top-down LiDAR view (rendered 11 meters above the vehicle, utilizing aforementioned ground segmentation and color grading).
This layout was designed to provide the operator with a global overview of the environment, supporting spatial orientation and obstacle localization, and also serving as a bridge between the abstract LiDAR data and the more intuitive camera feed. 
Furthermore, while previous work exploring this visualization approach reported that it not only fails to improve situational awareness but, in some cases, makes it more difficult for the teleoperator to perceive important features~\cite{46}, our interface was explicitly designed to challenge those findings in real world test.


\section{USER STUDY}
To evaluate the different information streams, we conducted a within-subject user study, the protocol of which was fully approved by the relevant Ethics Committee. The study was designed to examine how various types of displayed data affect both objective performance and subjective operator experience. Statistical analysis was performed using a conservative $\alpha = 0.05$, ensuring that only robust effects were considered significant. 

\subsection{Hypotheses}
Through our study, we sought to compare three different approaches to presentation of sensor data in full-scale vehicle teleoperation, i.e, (i) camera-based layout (Camera) (see Fig. \ref{view_fusion2}A), (ii) LiDAR-based layout (LiDAR) (see Fig. \ref{view_fusion2}B), as well as (iii) fusion of camera and LiDAR sensors (Camera+LiDAR) (see Fig. \ref{view_fusion2}C). Consequently, we investigated the following hypotheses:

\textbf{Hypothesis 1}: \textit{User Experience} $H_{01}$: Sensor data type does not affect usability or cognitive taskload. $H_{A1}$: Usability and cognitive taskload differ across sensor types.

\textbf{Hypothesis 2}: \textit{Performance} $H_{02}$: Teleoperation performance is independent of sensor type. $H_{A2}$: Teleoperation performance varies significantly by sensor type.





\subsection{Participants}
We recruited 18 participants using an opportunistic sampling method (age: \(M = 23.72, SD = 1.36\)). The participant pool consisted primarily of university students, who reported good vision, either naturally or corrected to normal, and except one, held a valid driver’s license. When reporting their familiarity with teleoperation using the Likert scale, only 5 participants reported having relevant experience (\textit{rather agree}). At the same time, the majority reported little or no experience (13/18). 
All participants signed a consent form before the study commenced.

\subsection{Experimental Design and Task}

For each participant, the entire experiment lasted approximately 45 minutes. The study began with a survey that collected basic demographic and background information. 

The next phase focused on familiarizing participants with the teleoperation platform and the testing circuit. To isolate the learning curve of the interfaces, we provided an extensive, non-time-limited familiarization phase focused solely on vehicle dynamics and course layout. By ensuring that participants first reached a self-reported level of comfort with the physical task, we minimized the effects of general task learning during the measured trials. The latter were administered in a fully balanced randomized order to ensure that any residual learning effect is distributed equally across all conditions, thereby neutralizing its impact on the results.

\begin{figure}[!t]
\centering
\includegraphics[width=2.5in]{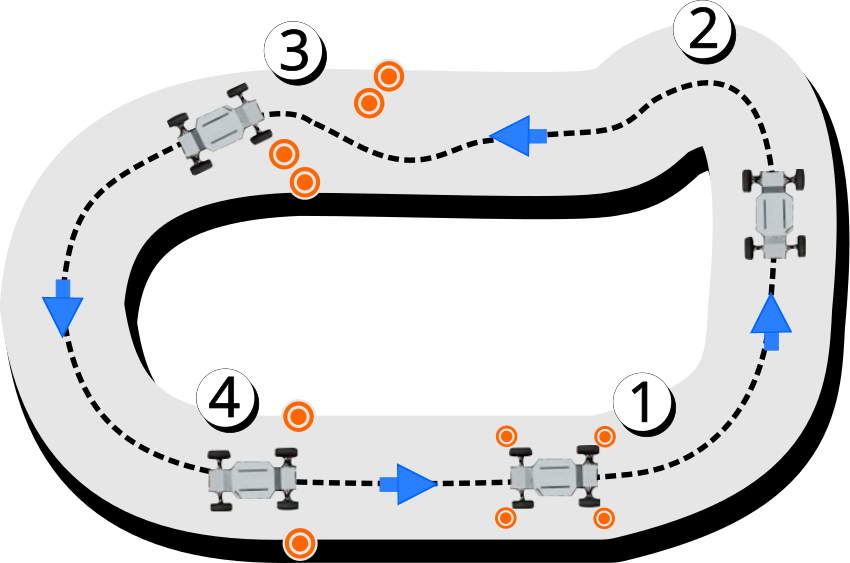}
\caption{The obstacle course with challenges in anti-clockwise direction: (1)~parking, (2)~tight turn, (3)~obstacle avoidance, and (4)~stop line.}
\label{fig_route}
\vspace{-2mm}
\end{figure}

\begin{figure}[!h]
\centering
\subfloat[Parking]{\includegraphics[width=0.48\linewidth]{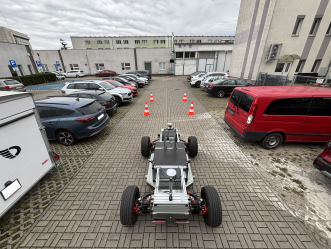}\label{fig:park}}
\hfill
\subfloat[Tight turn]{\includegraphics[width=0.48\linewidth]{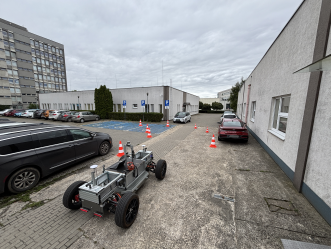}\label{fig:park}}\\[-0.2cm]
\subfloat[Obstacle avoidance]{\includegraphics[width=0.48\linewidth]{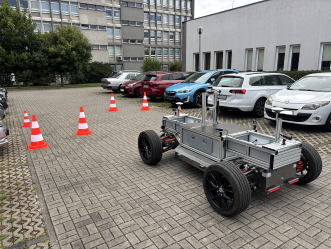}\label{fig:obs}}
\hfill
\subfloat[Stop line]{\includegraphics[width=0.48\linewidth]{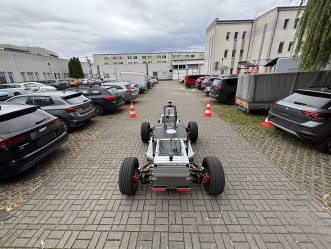}\label{fig:stop}}
\caption{Images presenting test circuit challenges in anti-clockwise direction: (1)~parking, (2)~tight turn, (3)~obstacle avoidance, and (4)~stop line.}
\label{fig_route2}
\vspace{-2mm}
\end{figure}

The test circuit measured approximately 40 meters in length and 10 meters in width (see Fig.~\ref{fig_route} and Fig. \ref{fig_route2}). The surface consisted primarily of paving stones, with a section of worn concrete. Several stationary vehicles were parked outside the circuit boundaries and remained in fixed positions throughout all participant trials. Moreover, our experiment was conducted under consistent meteorological conditions and within the same time window to ensure favorable lightning conditions. This real-life setting, with slight variations in pavement texture or ambient light, as well as the use of full-scale vehicles, significantly increased the ecological validity of our field study.

During the experiment, each participant completed the obstacle course three times, once with each proposed visualization. These were: (i) camera-based (see Fig. \ref{view_fusion2}A), (ii) LiDAR-based (see Fig. \ref{view_fusion2}B), and (iii) mixed Camera with LiDAR sensor fusion (see Fig. \ref{view_fusion2}C). To mitigate learning effects, the order of interface presentation was counterbalanced across participants using a Latin square design. To control for directional bias, the first half of users completed the course clockwise and the other half anticlockwise, ensuring that the relative order of obstacles remained unchanged. After each lap, participants completed the well-established \textit{System Usability Scale} (SUS) \cite{sus_org} and \textit{NASA Task Load Index} (NASA-TLX) \cite{Hart1988} questionnaires to evaluate perceived usability and workload, respectively.

Within each lap, participants performed a series of predefined tasks inspired by Georg et al.~\cite{48}, designed to simulate typical parking-lot scenarios and assess control precision, spatial awareness, and distance estimation:

\begin{enumerate}
\item \textit{Precision parking}:
Position the vehicle within a designated spot marked by cones and hold it stationary for approximately two seconds.
\item \textit{Tight turn}:
This task involved performing a tight turn within a space marked by cones, testing precise steering control and surroundings awareness.
\item \textit{Obstacle avoidance}:
Navigate a chicane to simulate maneuvering around dynamic or fixed obstacles.
\item \textit{Precision stopping}:
Stop the vehicle precisely at a marked finish line, allowing assessment of participants’ perception of the vehicle’s front boundary.
\end{enumerate}

The following experimental data were collected to establish objective metrics for performance estimation: (i) full lap time (\textit{time}), (ii) number of collisions with obstacles (\textit{collisions}), (iii) distance from the stop line  (\textit{stop}), (iv) distance from the center of the parking space (\textit{parking}, and (v) average error in position about the reference ride (\textit{trajectory}). To strictly isolate spatial deviation from variations in operator driving speeds, the \textit{trajectory} metric was defined as the average Euclidean distance between the vehicle's actual GNSS coordinates and the nearest points on a pre-recorded ground-truth (GT) reference path. These distances were efficiently computed using a nearest-neighbor search~\cite{NN}.




\section{USER STUDY RESULTS}

\subsection{Perceived Usability and Taskload}

To assess usability, participants completed the SUS survey after each lap. As the Shapiro-Wilk test did not detect any normality violations, we performed ANOVA, which revealed a statistically significant difference in SUS scores across the three conditions \((F(2,17) = 12.63, p=0.000035)\). 
Post-hoc Tukey’s tests confirmed the statistically significant difference across all three categories. Camera+LiDAR vs. LiDAR  ($F=26.806$,$p \leq 0.0001$, Cohen's $d=-1.806$), Camera vs. LiDAR ($F=13.611$, $p=0.036$, Cohen's $d=0.833$), and Camera+LiDAR vs. Camera ($F=13.194$. $p=0.043$, Cohen's $d=-0.788$).  
Notably, the sensor fusion BEV achieved a mean SUS score of 77.9/100, well above the 68 score widely considered the acceptability threshold~\cite{sus_org}, whereas the traditional Camera interface slightly underperformed relative to this benchmark, with LiDAR falling below this threshold~\cite{sus_org}.
This indicates that combining camera and LiDAR data in BEV sensor fusion improves perceived usability and reduces perceived taskload compared to the standard used in most modern teleoperating settings~\cite{52}.


\begin{figure}[!th]
\centering
\includegraphics[width=0.75\linewidth]{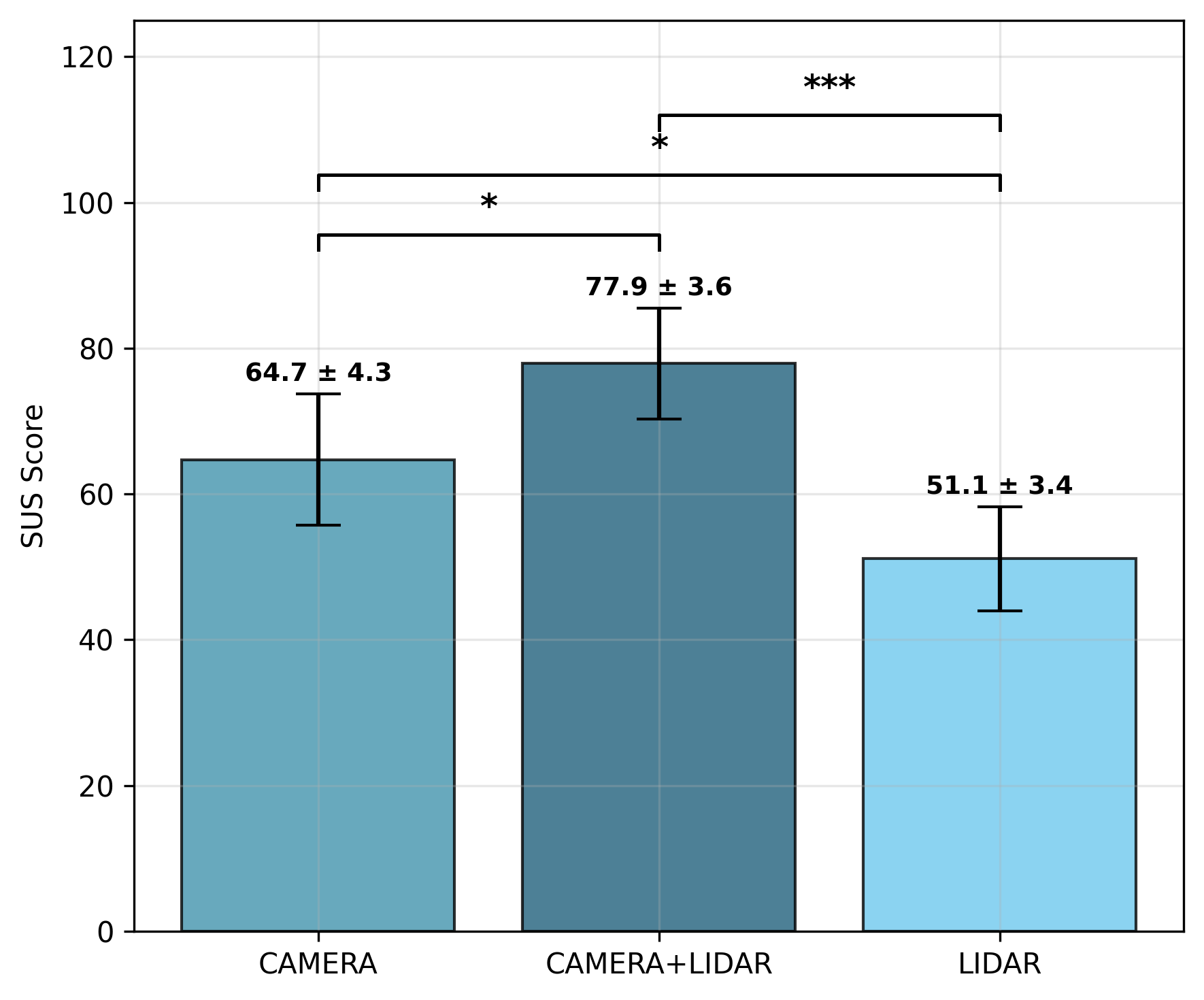}
\caption{Perceived usability scores measured with SUS for each interface ($N = 18$). Higher scores indicate better perceived usability. The whiskers denote standard error (SE) while asterisks denote statistical significance (one for \(p \leq 0.05\)) and three for \(p \leq 0.001\)).}
\label{fig_sus}
\vspace{-2mm}
\end{figure}

\begin{table}[ht]
\centering
\caption{SUS and NASA-TLX scores for each interface type \\
M = mean, SE = Standard Error} 
\begin{tabular}{lllllll}
\textit{View}         & \textit{Survey}  
                        & \emph{n}  & \emph{min}   & \emph{max}   & \emph{M}     & \emph{SE}   \\ \hline
\multirow{2}{*}{\textbf{Camera}}       & SUS$\uparrow$                     & 18 & 30.00  & 87.50  & 64.72 & 4.26  \\
                              & TLX$\downarrow$                     & 18 & 20.00 & 60.83 & 42.69 & 2.81 \\ \hline
\multirow{2}{*}{\textbf{Camera+LiDAR}} & SUS$\uparrow$                     & 18 & 47.50  & 100.0 & 77.92 & 3.61 \\
                              & TLX$\downarrow$                      & 18 & 5.00  & 55.83 & 34.58 & 3.05\\ \hline
\multirow{2}{*}{\textbf{LiDAR}}        & SUS$\uparrow$                     & 18 & 32.50  & 80.00  & 51.11 & 3.39 \\
                              & TLX$\downarrow$                     & 18 & 14.17 & 80.00 & 47.96 & 3.50
\end{tabular}%

\label{tab:tlxsus}
\vspace{-2mm}
\end{table}

\begin{figure}[ht]
    \centering
    \includegraphics[width=0.75\linewidth]{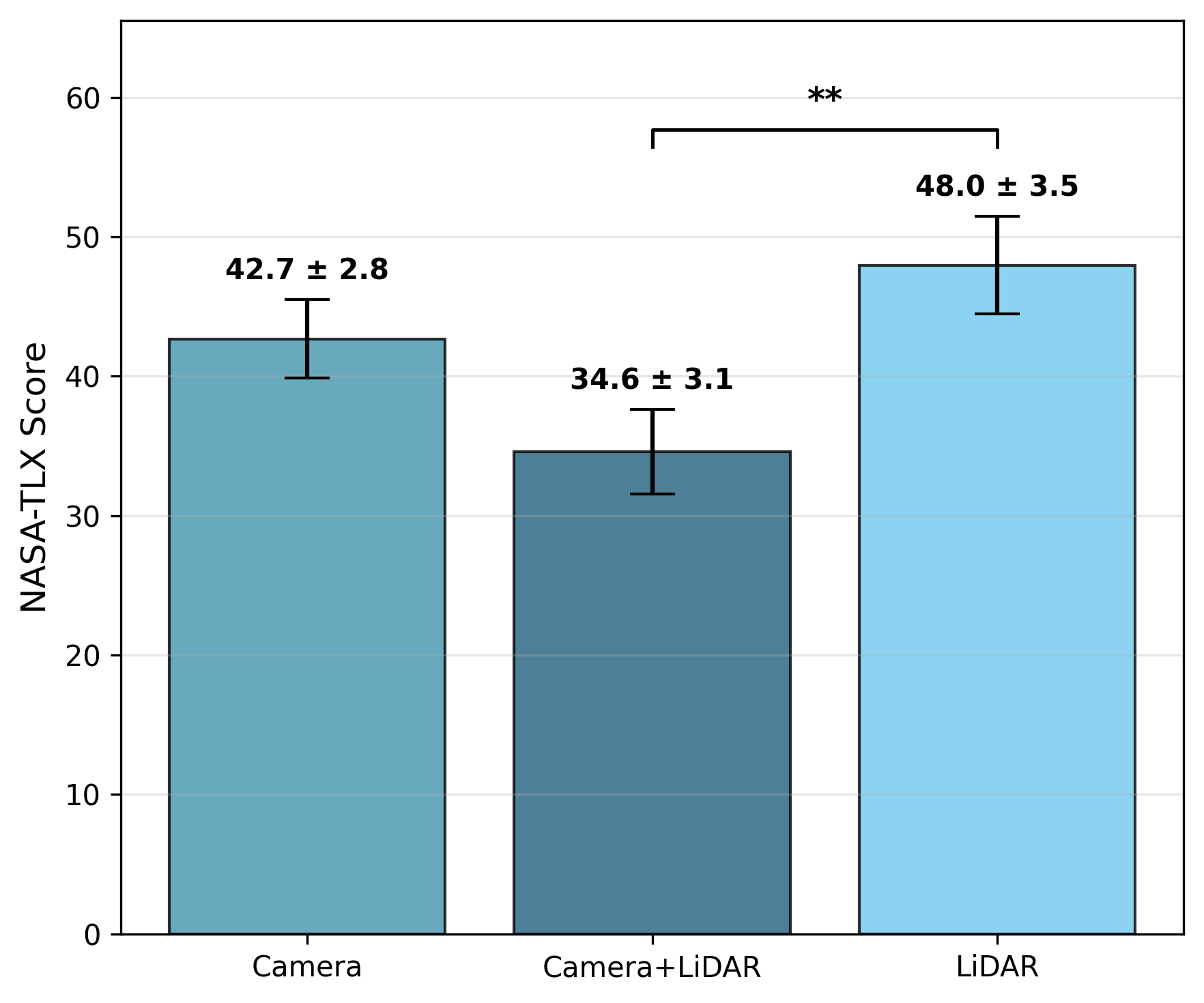}
    \caption{Perceived taskload levels measured with NASA-TLX showing that the multi-modal view of Camera+LiDAR was the least demanding in terms of workload experienced by the operator. The whiskers denote standard error (SE) while asterisks denote statistical significance (one for \(p \leq 0.01\)).}
    \label{fig:tlx1}
\end{figure}

To statistically evaluate differences in perceived taskload, NASA-TLX scores shown in Table~\ref{tab:tlxsus} were first tested for normality using the Shapiro-Wilk test, which revealed no significant deviations across conditions. A one-way ANOVA indicated a significant effect of interface type on perceived taskload \((F(2,17) = 4.626, p=0.014, \eta^2= 0.154)\). Post-hoc Tukey tests showed that the Camera+LiDAR interface differed significantly from the LiDAR-based interface ($F = 13.38$, $p \leq 0.01$ Cohen's $d=-0.961$), specifically, participants reported significantly lower task workload when using the Camera+LiDAR view compared to LiDAR alone \((M_{Diff} = -13.38, SE = 4.64, 95\%-CI[-22.82, -3.94])\), whereas other pairwise comparisons did not reach statistical significance (see Fig.~\ref{fig:tlx1}).

Consistent with the SUS findings, the NASA-TLX taskload assessment showed that the Camera+LiDAR interface was also associated with the lowest perceived workload (\(M = 34.58, SD = 12.93, 95\%-CI [28.15, 41.02]\)), demonstrating that fusion reduces mental effort despite presenting multiple data sources. The Camera view produced intermediate workload scores (\(M = 42.69, SD = 11.91, 95\% CI [36.76, 48.61]\)), suggesting moderate cognitive demand. Conversely, the LiDAR view resulted in the highest workload scores (\(M = 47.96, SD = 14.84, 95\%-CI [40.58, 55.35]\)), highlighting the cognitive strain caused by presenting operators with a representation that departs significantly from natural visual experience. 

Based on these findings, we reject the \textit{null hypothesis} $H_{01}$ and support the \textit{alternative hypothesis} $H_{A1}$ that there is an observable and statistically significant difference between Camera, LiDAR, and Camera+LiDAR in terms of usability and cognitive taskload, with the latter condition resulting in the significantly higher usability and requiring smaller taskload than LiDAR alone. It was also significantly better than the mean SUS score under the Camera condition, which did not cross the acceptable usability threshold.


\subsection{Teleoperation Performance}

\begin{table*}[!ht]
\centering
\caption{Obtained quantitative metrics after filling missing data and removing outliers}
\begin{tabular}{llcclc}
  \textit{View}           & \textit{Time [s]}             & \textit{No. of collisions}           & \textit{Stop distance [cm]}           & \textit{Parking [cm]}          & \textit{Mean Trajectory Deviation [m]}    \\ \hline
\textbf{Camera}       & 149.2 ± 9.14  & 1.529 ± .259 & 44.8 ± 7.8 & 152.3 ± 18.3  & 0.665 ± .082 \\ 
\textbf{Camera+LiDAR} & 144.0 ± 10.2  & 1.647 ± .383 & 26.7 ± 6.3 & 145.2 ± 21.1  & 0.635 ± .063 \\ 
\textbf{LiDAR}        & 167.8 ± 9.65  & 1.875 ± .340 & 27.9 ± 6.1 & 186.5 ± 27.5  & 0.830 ± .090
\end{tabular}%

\label{tab:obj_results}
\end{table*}



\begin{figure*}[!ht]
    \centering
    \includegraphics[width=1.0\linewidth]{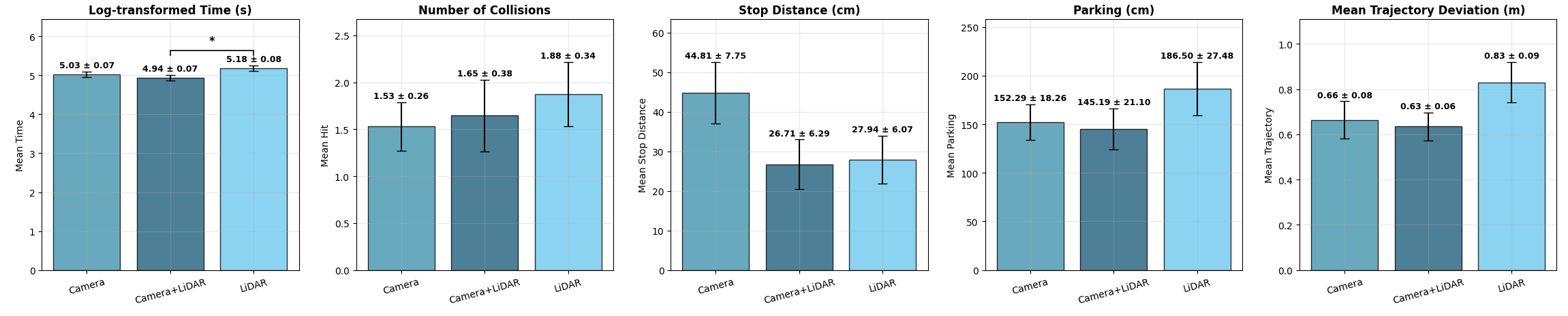}
    \caption{Results from the objective data, in their original form, indicate that lower values correspond to better performance. In this form, a clear pattern emerges: the fused BEV enables users to drive more quickly, with fewer collisions and greater accuracy.}
    \label{fig:objective}
    \vspace{-3mm}
\end{figure*}

All quantitative data collected during the experiment were subjected to post-processing before statistical analysis (see Table~\ref{tab:obj_results}). The missing values were responsible for approximately 8.54\% of the dataset. Outliers were identified and removed using the interquartile range (IQR). This procedure resulted in the removal of approximately 7.42\% anomalous data points, ensuring that extreme, unrepresentative maneuvers did not skew the subsequent statistical tests.
Due to the discrete nature of \textit{collisions} data, it was analyzed separately using Generalized Linear Model (GLM) Regression. Pairwise contrasts showed that none of the comparisons within the \textit{collisions} category were statistically significant.


For \textit{parking}, the Shapiro–Wilk test indicated no significant deviation from normality; hence, we applied a one-way ANOVA for analysis. In contrast, \textit{time}, \textit{trajectory}, and \textit{stop} did not meet normality assumptions. Therefore, we used the non-parametric Kruskal–Wallis test for their analysis. To reduce skewness, \textit{time} data were log-transformed~\cite{log_trans}.

We observed a statistically significant effect of \textit{time} ($F=6.0854, p=0.0477$). Analysis with Dunn’s post hoc test, adjusted with Holm-Bonferroni correction, confirmed a significant difference between the LiDAR and Camera+LiDAR views ($p=0.0461$, Cohen's $d=0.830$) while no other pairwise comparisons reached significance (LiDAR vs. Camera: $p = 0.438$, Cohen's $d=-0.516$); Camera vs. Camera+LiDAR: $p = 0.208$, Cohen's $d=0.324$).


Based on these findings, we partially support the \textit{alternative hypothesis} $H_{A2}$ that there is an observable difference in teleoperation performance between the conditions. Specifically, between the LiDAR BEV and Camera+LiDAR BEV conditions, the latter yields shorter task completion times. At the same time, enhancing the Camera with LiDAR data did not yield completion times significantly shorter than those with Camera data alone. 

\section{Discussion}

Our results challenge the assumption that multi-modal fusion would increase cognitive load due to perspective-merging requirements~\cite{46}. Instead, objective performance metrics—including task completion time ($p=0.0477$) align with the subjective SUS ($77.9/100$) and NASA-TLX ($34.58/100$) trends. This synergy suggests that Camera+LiDAR fusion effectively mitigates the ``sensory gap'' inherent in remote operation~\cite{52, being} by grounding abstract spatial data within an intuitive visual context~\cite{64}.

A notable deviation occurred in stopping accuracy, where the standalone LiDAR interface and BEV fusion view provided greater precision. We attribute this to the isometric perspective of the LiDAR BEV, which eliminates focal-length distortions and depth compression typical of monocular camera feeds~\cite{bevreview}, allowing for more precise ego-vehicle localization relative to the stop line. However, this spatial advantage did not translate into improved general navigation; the LiDAR-only condition yielded the highest collision rates and the greatest trajectory deviation. This suggests that while LiDAR excels at discrete distance estimation, it lacks the semantic richness required for situational awareness. This modality was intentionally represented in its raw form as a controlled baseline to isolate the specific impact of abstract depth perception from the semantic richness inherent in camera-based feeds.

Furthermore, participant feedback emphasized that co-located sensor placement is critical for minimizing the "transformation penalty" during perspective shifts. Our findings suggest that the efficacy of a teleoperation interface is not determined by the volume of data, but by the degree to which the representation synchronizes depth precision with semantic clarity. By optimizing this trade-off, the fused BEV interface maximizes safety margins while significantly reducing the operator's mental workload.

\section{Conclusion}
Our study demonstrates that a multi-modal Camera+LiDAR interface improves teleoperation performance and operator experience compared to single-sensor setups. The fused interface led to statistically significant reductions in task completion time relative to LiDAR-only BEV and to superior SUS and lower cognitive workload scores. These findings suggest that BEV fusion enhances situational awareness by grounding abstract point cloud data in an intuitive camera-based context, whereas the standalone LiDAR interface proved difficult to interpret for effective maneuvering. At the same time, Camera-only view barely approached the perceived acceptable usability threshold \cite{sus2}, although with similar performance as fusion BEV.
Moving forward, our research will focus on making the fused Camera+LiDAR BEV visualizations more human-readable by projecting camera textures directly onto 3D point clouds and utilizing instance segmentation to classify environmental obstacles. Additionally, in our future work, we will conduct longitudinal studies with a diverse demographic participant pool to provide a broader perspective on long-term adaptation.

\bibliographystyle{ieeetr}
\balance
\bibliography{Bibliography}

%








\end{document}